\documentclass[conference, a4paper]{IEEEtran}
\IEEEoverridecommandlockouts

\ifCLASSINFOpdf
  \usepackage[pdftex]{graphicx}
  \DeclareGraphicsExtensions{.pdf,.jpeg,.png}
\else
\fi

\usepackage{cite}
\usepackage{amsmath,amssymb,amsfonts}
\usepackage{graphicx}
\usepackage{booktabs}
\usepackage{url}

\usepackage{algorithmic}
\usepackage[dvipsnames]{xcolor}

\usepackage{subcaption}
\usepackage{multirow}
\usepackage{array}
\makeatletter
\newcommand*\titleheader[1]{\gdef\@titleheader{#1}}
\AtBeginDocument{%
\let\st@red@title\@title
\def\@title{%
\bgroup\normalfont\normalsize\centering\@titleheader\par\egroup
\vskip0.2em\st@red@title}
}
\makeatother

\makeatletter
\renewcommand{\fnum@figure}{Figure \thefigure}
\makeatother

\title{ {Efficient Real-Time Aircraft ETA Prediction via
Feature Tokenization Transformer} \\
\thanks{This work is supported by the National Research Foundation, Singapore, and the Civil Aviation Authority of Singapore (CAAS), under the Aviation Transformation Programme. (Grant No. ATP2.0\_WIC\_I2R for ATM\_I2R\_2).}
\vspace{0.5cm}
}

\titleheader{First US-Europe Air Transportation Research and Development Symposium (ATRDS2025)}

\author{\IEEEauthorblockN{Liping Huang, Yicheng Zhang, Yifang Yin, Sheng Zhang, Yi Zhang}
\IEEEauthorblockA{Institute for Infocomm Research \\
Agency for Science, Technology and Research \\
Singapore \\
}
}

\renewcommand\IEEEkeywordsname{Keywords}
\IEEEaftertitletext{\vspace{-1\baselineskip}}

\begin{document}

\maketitle

\noindent \begin{abstract}
Estimated time of arrival (ETA) for airborne aircraft in real-time is crucial for arrival management in aviation, particularly for runway sequencing. Given the rapidly changing airspace context, the ETA prediction efficiency is as important as its accuracy in a real-time arrival aircraft management system. In this study, we utilize a feature tokenization-based Transformer model to efficiently predict aircraft ETA. Feature tokenization projects raw inputs to latent spaces, while the multi-head self-attention mechanism in the Transformer captures important aspects of the projections, alleviating the need for complex feature engineering. Moreover, the Transformer's parallel computation capability allows it to handle ETA requests at a high frequency, i.e., 1HZ, which is essential for a real-time arrival management system. The model inputs include raw data, such as aircraft latitude, longitude, ground speed, theta degree for the airport, day and hour from track data, the weather context, and aircraft wake turbulence category. With a data sampling rate of 1HZ, the ETA prediction is updated every second. We apply the proposed aircraft ETA prediction approach to Singapore Changi Airport (ICAO Code: WSSS) using one-month Automatic Dependent Surveillance-Broadcast (ADS-B) data from October 1 to October 31, 2022. In the experimental evaluation, the ETA modeling covers all aircraft within a range of 10NM to 300NM from WSSS. The results show that our proposed method method outperforms the commonly used boosting tree based model, improving accuracy by 7\% compared to XGBoost, while requiring only 39\% of its computing time. Experimental results also indicate that, with 40 aircraft in the airspace at a given timestamp, the ETA inference time is only 51.7 microseconds, making it promising for real-time arrival management systems. 
\end{abstract}

\vspace{0.3cm}

\begin{IEEEkeywords}
Estimated Time of Arrival; Tokenizer; Transformer; Arrival Aircraft Management; Data-Driven Simulation
\end{IEEEkeywords}

\section{Introduction (\textit{Heading 1})}
The aircraft estimated time of arrival (ETA) is the time when an aircraft is expected to touch down on the runway \cite{ct01}. Accurate ETA predictions are essential for air traffic controllers to manage aircraft flow more effectively, providing curcial information on the time to lose and time to gain \cite{c02, c05}. This ensures airspace safety and reduces congestion and delays \cite{c08, c09}. Due to the rapidly changing airspace context, real-time ETA predictions play a pivotal role in enhancing air traffic management. They optimize runway sequencing and dynamically provide air traffic controllers with the necessary adjustments of the time to lose or gain.  As a vital component of the arrival management system in aviation, the efficiency and effectiveness of aircraft ETA prediction are both paramount \cite{ct02, c06}. Accurate predictions help optimize resources and minimize delays, while efficient predictions adapt to varying contexts and ensure smoothing operations.

The prediction of aircraft ETA has evolved significantly, with various methods being explored in the literature. Traditional approaches often relied on deterministic models, which use aircraft performance data and physics-based trajectory models to estimate arrival times \cite{ct03}. While foundational, these methods can struggle with the dynamic nature of air traffic and weather conditions.  More recent advancements have integrated machine learning techniques. For example, random forests and deep neural networks have been employed to predict aircraft ETA by learning from historical flight data and weather conditions \cite{c017}. An advantage of machine learning-based models is their ability to adapt to new data and provide more accurate predictions compared to traditional methods \cite{ct04}.

The published methods in the literature are primarily designed for one-time aircraft ETA predictions. However, for an online arrival management system, ETA predictions need to be frequently updated due to the dynamically changing airspace context. These predictions should be made in a rolling-horizon manner to capture contextual changes \cite{ct00}, thereby providing necessary adjustments to air traffic controllers. Despite advancements of utilizing machine learning method in aircraft ETA predictions, real-time ETA prediction remains a challenging area. Further research and development are needed to enhance the  efficiency of these predictions in real-time scenarios \cite{ct04, ct00}.

\begin{figure*}[tbh]
    \centering
    \begin{subfigure}[b]{0.455\textwidth} 
        \centering
        \includegraphics[width=\textwidth]{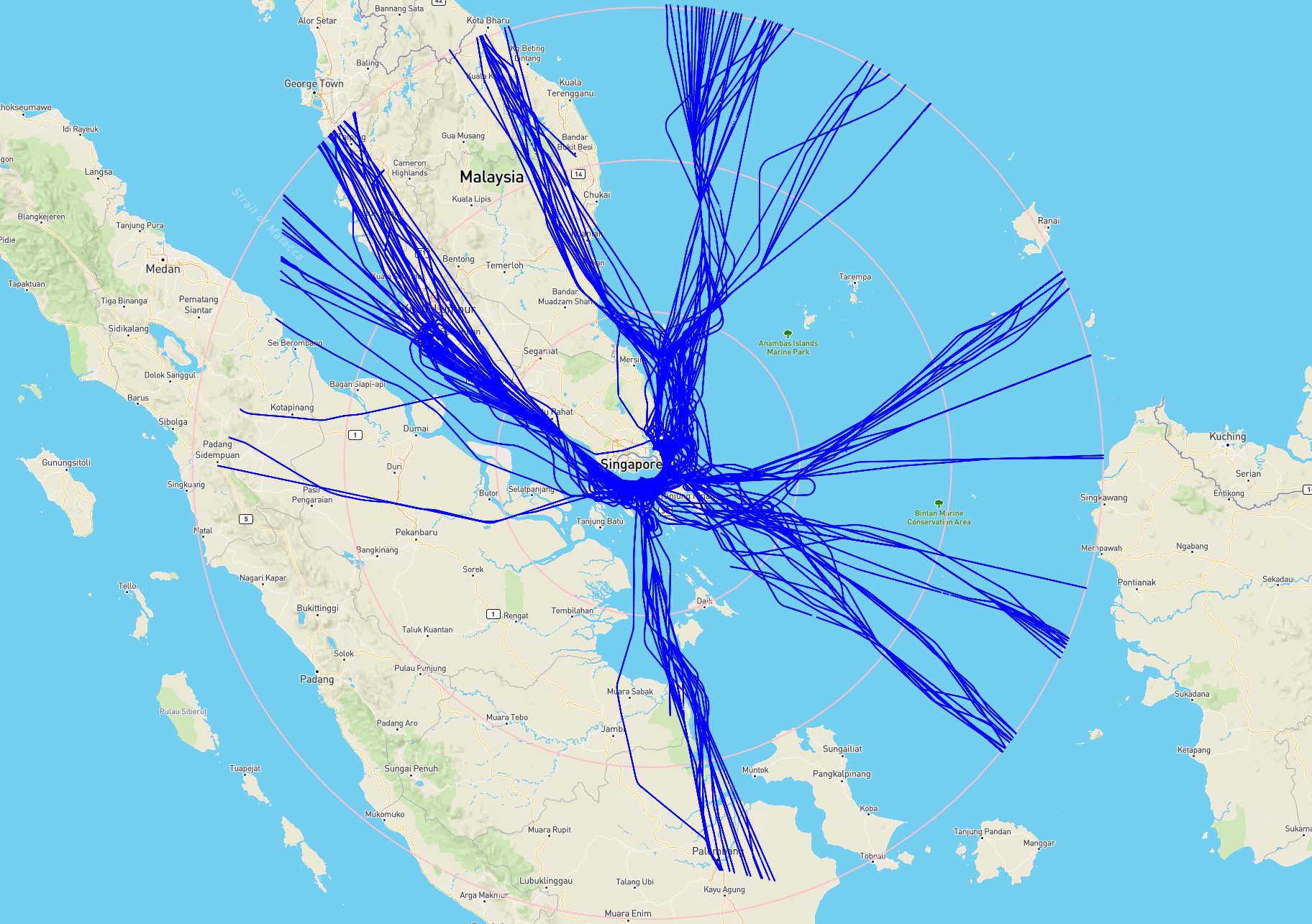} 
        \caption{Trajectories of Arrival Aircraft within a Single Day from 00:00AM to 01:00 PM. Pink Circles denotes the 100NM, 200NM, 300NM from WSSS.}
        \label{sub1}
    \end{subfigure}
    \hspace{0.01\textwidth}
    \begin{subfigure}[b]{0.46\textwidth} 
        \centering
        \includegraphics[width=\textwidth]{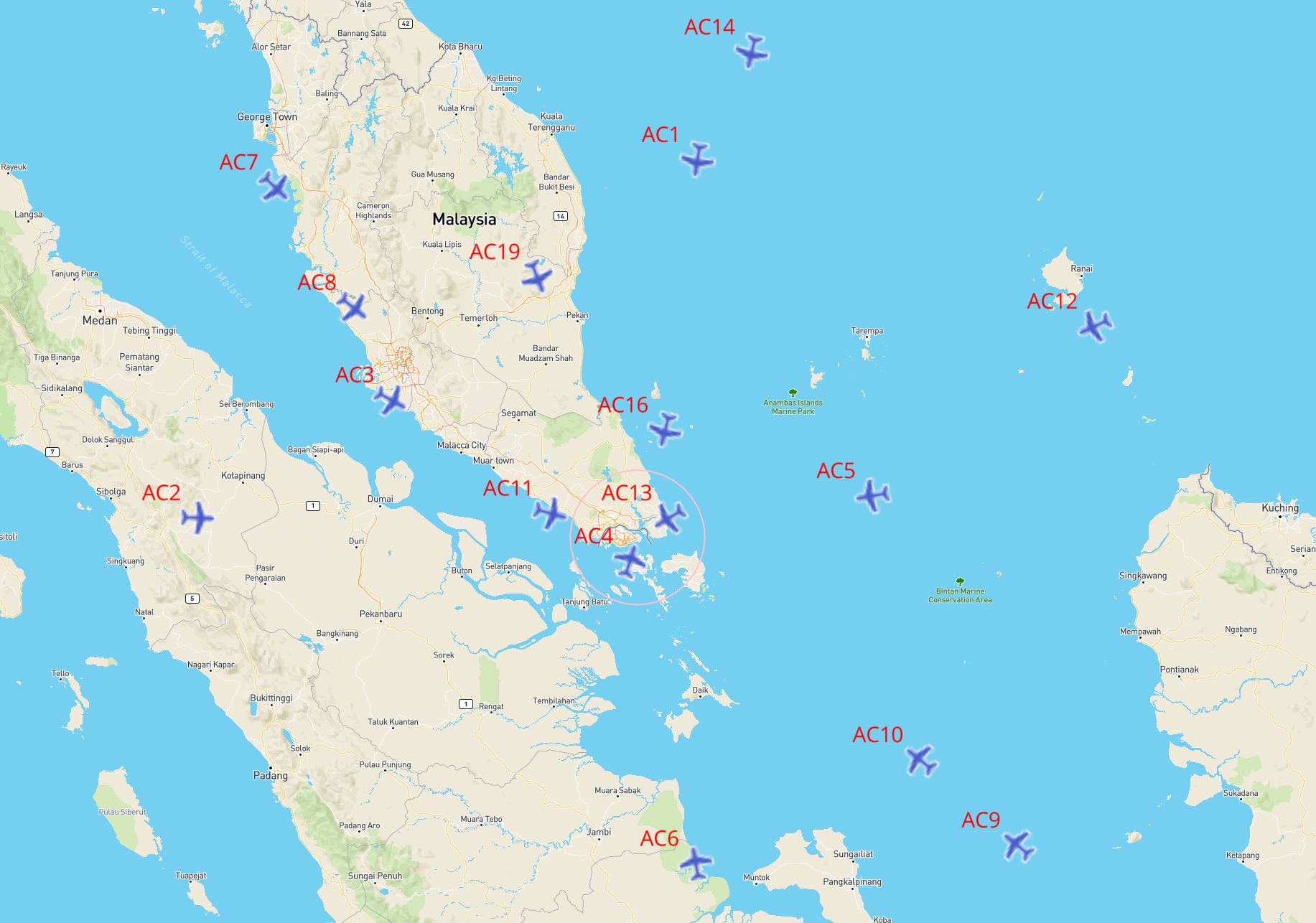} 
        \caption{A Snapshot of Airborne Arrival Aircraft for WSSS. The Pink Circle is 40NM from WSSS, Which is the TMA Boundary of WSSS. }
        \label{sub2}
    \end{subfigure}
    \caption{Aircraft Trajectories and A Snapshot Example of Airborne Arrival Aircraft. The ETA modeling method utilizes the aircraft trajectories as inputs and the prediction can be for all airborne arrival aircraft as in (b).}
    \label{fig1}
\end{figure*}

As shown in Fig. \ref{fig1}, we propose utilizing aircraft trajectory data to model the ETAs of all airborne arrival aircraft for an airport. Fig. \ref{sub1} illustrates the trajectories of all aircraft landing at WSSS within a single day from 0:00 AM to 01:00 PM. For the prediction phase in the real-time arrival management system, the model predicts ETAs for all airborne aircraft, as depicted in the snapshot in Fig \ref{sub2}. This real-time predictions can help sequence arrival aircraft for runway usage. Moreover, we propose predicting the ETA at a high frequency of 1HZ for he real-time arrival management. This means that when we utilize the prediction model in the arrival management system, we obtain a snapshot like Fig. \ref{sub2} every second. 

We propose utilizing the parallel computation characteristics of the Transformer neural network \cite{ct000} for real-time aircraft ETA predictions, which is not explored in existing literature. The transformer neural network leverages self-attention mechanisms, enabling more efficient parallel processing compared to traditional recurrent neural networks. The architecture significantly reduces the computational complexity and training time. Furthermore, we employ feature tokenization to create a data-light aircraft ETA prediction model, where simple raw data inputs are automatically projected to latent spaces and further fed into the Transformer network, eliminating the need for complex feature engineering.

In the case study of Changi Airport presented in this paper, we propose to predict the ETA of aircraft up to 300NM from the airport. The data sampling rate and the ETA prediction frequency are both 1HZ, For each given timestamp, the ETA prediction covers all aircraft within the 300NM airspace of the airport. Experimental evaluation shows that for each second's data (approximately 40 aircraft in the airspace), the ETA inference time is only 57.1 microseconds. This demonstrates a high potential for utilization in real-time arrival management systems. 
Contributions of this study are summarized as follows:
\begin{itemize}
    \item We propose adopting the transformer neural network for real-time aircraft ETA prediction. This approach fully leverages the parallel computation characteristics of the Transformer to achieve efficient ETA prediction, making it highly suitable for real-time arrival management systems in aviation. 
    
    \item To simplify the feature engineering process, we propose using feature tokenization on raw data inputs. This method projects the raw input features into high-dimensional latent spaces to generate learnable feature embeddings. These embeddings are then fed into the Transformer neural networks, and he multi-head self-attention block automatically captures the important aspects of the input feature embeddings, and outputs the final ETA predictions. 
    
    \item The proposed method has been validated through a case study of Singapore Changi Airport, using one-month ADS-B data from October 1 to October 31, 2022. Experimental results show that our method improve prediction accuracy compared to the commonly used tree-based XGBoost model, while requiring less training time. Our analysis also indicates the inference efficiency for real-time trajectory data, demonstrating its potential for use in real-time arrival management systems.
\end{itemize}

The rest of this paper is organized as follows. Section \ref{relatedwork} summarizes the related work. Section \ref{Method} presents the methodology. Section \ref{experiment} presents the experimental evaluations and corresponding analysis of the prediction accuracy. Conclusions and discussion for future work are finally drawn in Section \ref{conclusion}.

\begin{figure*}[ht]
\begin{center}
	\includegraphics[width=0.85\textwidth]{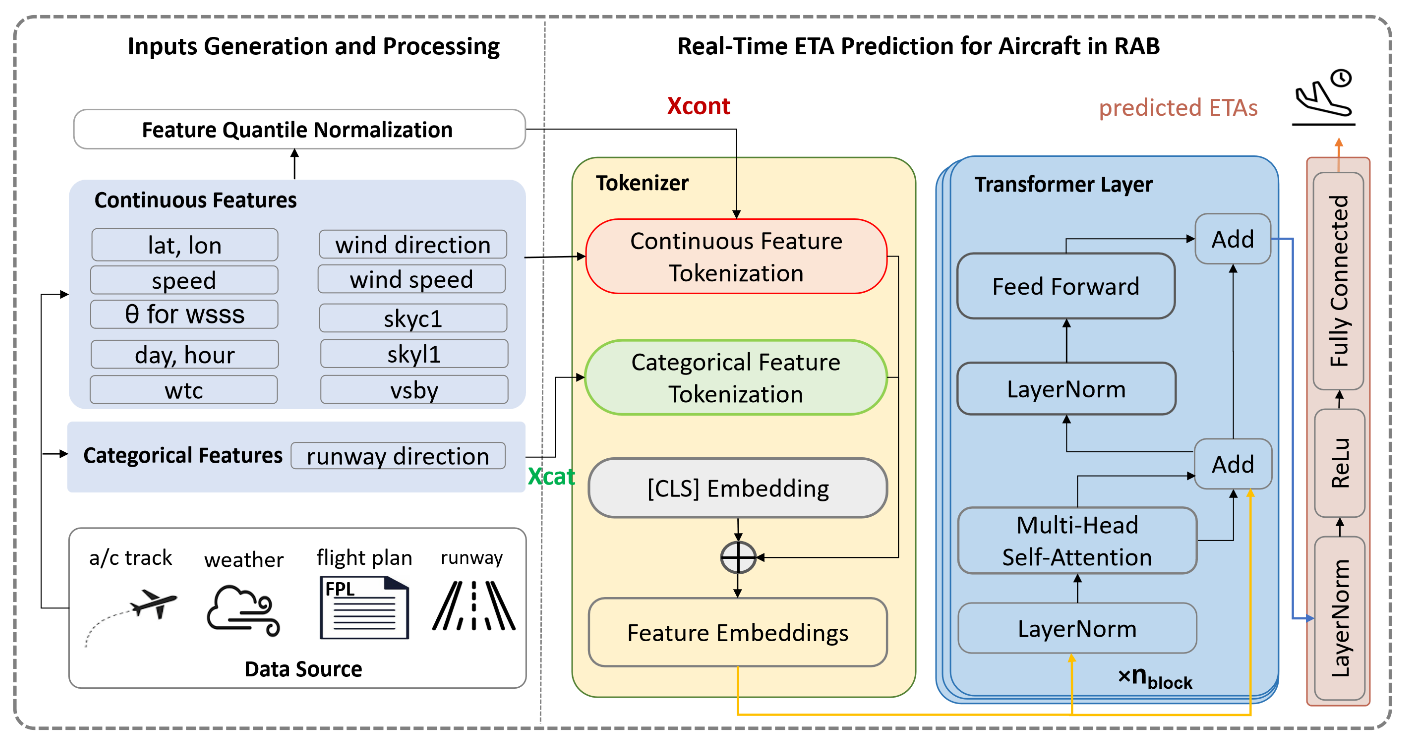}
	\caption{The Proposed ETA Prediction Model. The Raw Inputs are from The ASD-B Track Data, and the METAR Weather Condition Data. The Runway Direction is Set as Categorical Input, and All Other Inputs are Set as Continuous Number Inputs. }
	\label{framework}
 \end{center}
\end{figure*}

\section{Related Work}
\label{relatedwork}
Given the importance of aircraft ETA prediction, various data-driven machine learning-based approaches have been developed in the literature. The most commonly used data is the aircraft track data since it involves the aircraft positioning information, which is critical for ETA prediction \cite{ct03, ct05}. Specifically, the latitude, longitude and altitude from track data are directly used in the ETA modeling. Further derived features, such as distance to runway \cite{ct02, c007}, speed \cite{c004, ct05, c013}, climb rate \cite{c006 }, heading \cite{c006}, are also utilized. Additionally, the temporal information such as the time of day, day of week, and meteorological context are incorporated into the ETA modeling \cite{c006}. The study in \cite{c005} adopts the latitude and longitude to generated trajectory images for capturing complex airspace traffic context. 

Meanwhile, further calculated features, such as the traffic density \cite{ct03, c012, c015}, congestion degree and trajectory patterns \cite{c007} are adopted in the ETA modeling. The research in \cite{ct02, ct05} also incorporates the STAR route for the ETA modeling. However, the map matching of trajectories to STAR can be challenging, as aircraft do not always follow the STAR route precisely due to traffic control. The study in \cite{c014} explores additional features from the trajectories for ETA modeling, including the calculated arrival pressure and the sequencing pressure. Th extensive data requirements and complex calculations in these methods make them less adaptable to real-time changes in flight conditions.

In terms of the prediction modeling methods, the most commonly applied machine learning techniques are the decision tree-based models \cite{c015, c016}. The performance of these kind of models heavily depends on the features, requiring extensive feature engineering work \cite{c017, c018}. Conversely, deep learning-based methods can help capture the unseen patterns in the data. The research in \cite{neural} proposes a deep learning-based method for ETA prediction, where the the time series-based deep learning model, LSTM, is used. However, the recurrent computation mechanisms of LSTM are inefficient as they cannot be processing in parallel like the Transformer neural network. The parallelism of Transformer significantly reduces training time and computational complexity. Additionally, the self-attention mechanism in Transformers allows it to focus on relevant parts of the input data, enhancing their performance and scalability \cite{c00}. These features make Transformers particularly well-suited for real-time computation tasks and capturing complex contexts hidden in data.

\section{Methodology}
\label{Method}
\subsection{Problem Statement}

\begin{table*}[h]
  \begin{center}
    \caption{Acronym (Variable) and Corresponding Definition}
    \label{tab_acronyms}
    \begin{tabular}{|c|c|c|}
    \hline
      \textbf{Acronym (Variable)} &\textbf{Definition} & \textbf{Description} \\
      \hline
      OAB & Outer Airspace Boundary  & 300 NM to Changi Airport\\
      \hline
      IAB & Inner Airspace & 10 NM to Changi Airport\\
      \hline
      RAB & Research Airspace between OAB and IAB & the 2D airspace between OAB and IAB\\
      \hline
     $t$ & a given timestamp referencing to current time  & all information after this timestamp is unknow\\
      \hline
      $\mathbf{t}_{j}^{L}$ & timestamp when aircraft $j$ touches down on the runway  & this is extracted by matching track data to the runway\\
      \hline
      $\mathbf{T}_j^t$ & ETA for aircraft $j$ at time $t$, calculated as $\mathbf{t}_{j}^{L} - t$& target prediction variable\\
      \hline
      $X_j^{t}$ & feature inputs at timestamp $t$ for aircraft $j$ & used to represent information that can affect $\mathbf{T}_j^t$\\
      \hline
    \end{tabular}
  \end{center}
\end{table*}

The acronyms, variables, and corresponding definition used in this paper are summarized in Table \ref{tab_acronyms}. In this study, we propose to predict the landing time of each inbound aircraft that are in the RAB (10NM to 300 NM) of Singapore Changi Airport. The ETA truth label for aircraft $j$ at a given timestamp $t$ is $\mathbf{T}_j^t$, which is calculated as as $\mathbf{T}_j^t=\mathbf{t}_{j}^{L} - t$, where $t$ is the given timestamp at which the the ETA prediction for aircraft $j$ is conducted, and $\mathbf{t}^{L}_{j}$ is the truth timestamp when aircraft $j$ touches down on the runway threshold, which is extracted by matching the historical track data to the runway of Changi Airport. For predicting $\mathbf{T}_j^t$, we intend to construct a mapping function $\textbf{T}_j^t=\mathcal{F}(X_j^t)$, where $X_j^t$ is the feature inputs that capture the factors of ETA modeling at time $t$ for aircraft $j$. In this study, we utilize the Transformer Neural Network to construct the mapping function, and the input feature $X_j^{t}$ is a combination of positioning, temporal and meteorological information, which will be described in the later section. Hence, the objective is to construct the mapping function as 

\begin{equation}
    \mathbf{T}_{j}^t = \mathcal{F}(X_{j}^t), j\in\mathcal{J}_t, t\in\mathcal{T}
    \label{target}
\end{equation}
where $\mathcal{J}_t$ is the set of all aircraft in the dataset in the training phase. $\mathcal{T}$ is all the timestamp set, e.g., for a one-day dateset and data sampling is one point per second, then timestamp set $\mathcal{T}$ is all seconds in this day. In the inference phase of utilizing the model in \eqref{target} for real-time ETA prediction, the aircraft set $\mathcal{J}_t$ is all the airborne aircraft in RAB, as illustrated in Fig. \ref{sub2}.

Data sampling is one point per second in our case study of this paper, hence the ETA prediction frequency is 1HZ. The high prediction frequency allows for real-time response of ETA request for airport arrival management. In the following subsections, we elaborate on the methodology, data source, data analysis, and summary of the input feature of the model.

\subsection{ETA Prediction Model}
\subsubsection{Model Architecture}
The framework for the proposed real-time aircraft ETA prediction method is illustrated in Fig. \ref{framework}. The model's data sources include aircraft track data, weather information, flight plan data, and runway configuration. Both continuous and categorical features are extracted from these data sources. All continuous features undergo quantile normalization. The tokenizer model processes these features, producing feature embeddings. The Transformer layer then utilizes these feature embeddings to generate real-time ETA predictions. The transformation from the inputs to output embeddings are learnable within the model via the gradient descent algorithm. Further details on the tokenization and the revised version of Transformer for ETA prediction are provided in the subsequent subsections.

\subsubsection{Quantile Normalization} For the input data with $X^{m\times n}$, where $m$ is the number of samples, and $n$ is the number of features, the quantile normalization algorithm sort the values for each feature, and determines their ranks. Then it computes the quantiles for each feature, which involves creating a mapping from each rank to its corresponding value in the sorted array. Each original value in the dataset is replaced by the value corresponding to its rank in the sorted list. This means that if a value has a rank of $r$, it will be replaced by the value at the $r^{th}$ quantile of the featgure's distribution. In experiments, we set the output distribution as normal distribution. So, in this study, we normalize the data by transforming it based on the ranks of the values, ensuring that the resulting distribution is the normal across all samples. This transformation can significantly enhance the performance of machine learning models by making the data more comparable.

\subsubsection{Feature Tokenization}
\begin{figure}[ht]
\begin{center}
	\includegraphics[width=0.4\textwidth]{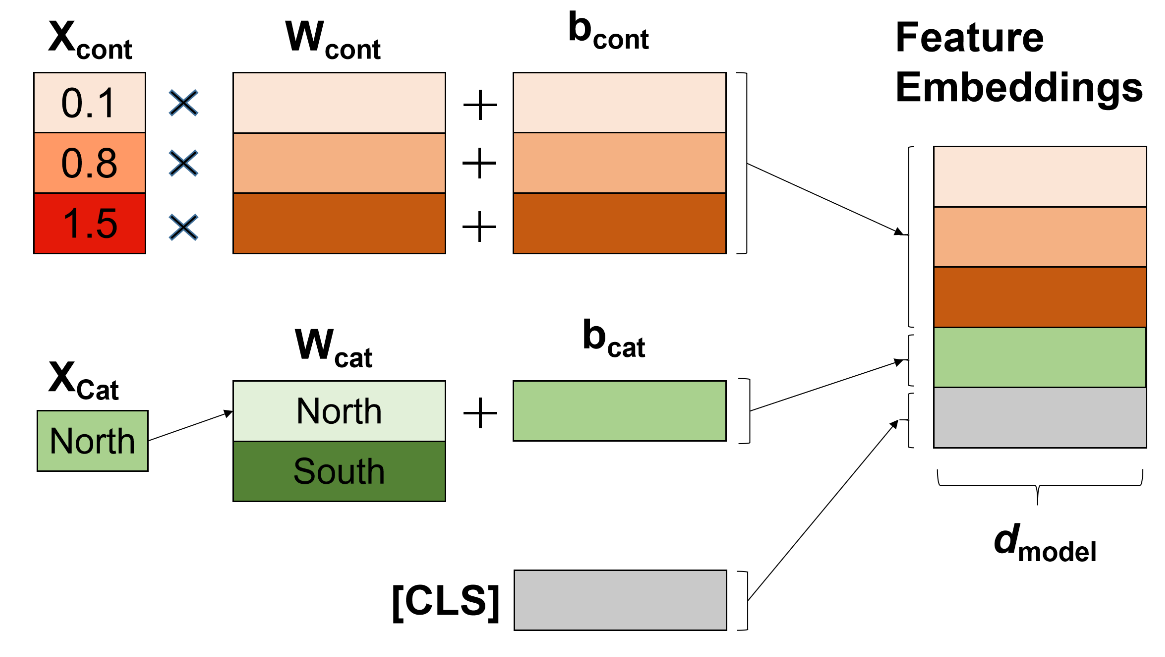}
	\caption{An illustration for the Feature Tokenization. The Parameters $W_{cont}, b_{cont}, W_{cat}, b_{cat}$ and the class token [CLS] Are Learnable. $d_{model}$ is the Hyperparameter of the Model. }
	\label{ft}
 \end{center}
\end{figure}

The feature tokenizer transforms the input features $\mathbf{x}\in\mathbb{R}^{k}$ to embeddings $X^E\in\mathbb{R}^{k\times d}$ \cite{c00}. The embedding for a specific feature $\mathbf{x}_i$ is computed as $X^E_i=b_i+f_i(x_i)\in\mathbb{R}^d$. As shown in Fig. \ref{ft},  $f_i$ is the linear projection function represented as the projection weight $W_i^{num}$. For the categorical features, the projection function is implemented as the lookup table $W_i^{cat}$. For example, if the categorical variable $x_{cat}$ has two categorical values, such as North and South in Fig. \ref{ft}, then the projection matrix dimension is $W^{cat}\in\mathbb{R}^{2\times d}$. Similarly, if the categorical variable $x_cat$ has three category values, the projection matrix is $W\in\mathbb{R}^{3\times d}$. Here $d$ is the model dimension, which is a model hyperparameter.

\subsubsection{Transformer}
Compared to Encoder in the original Transformer architecture, a pre-normalization is added before the Multi-Head Self-Attention module to facilitate easier optimization. Additionally, a [CLS] token is stacked with the continuous feature embedding $X_{cont}$ and categorical feature embedding $X_{cat}$, forming $X=stack[[CLS], X_{cont}, X_{cat}]$. The LayerNorm($X$) is then passed to the multi-head self-attention block. Specifically, the multi-head self-attention module partitions the inputs into $H$ parts, where $H$ is the number of head, and each part undergoes self-attention calculation. The self-attention mechanism maps the normalized feature embeddings to three separate space $Q, K, V$ using learnable linear transformations $W^Q, W^K, W^V$. The output is calculated as $softmax(\frac{Q\times K^T}{\sqrt{d_k}})\times V$. These $H$ outputs are further concatenated and feed into a linear projection as $Multi\-Head (Q, K, V)=Concat(head_1, head_H)W^O$, where $head_i = SelfAttention(QW^Q_i, KW^K_i, VW^V_i)$. In this study, the number of head is set as the default value of 8 as in Transformer. 


\subsection{Data Source}

\subsubsection{Aircraft Track Data}
The aircraft track data source of this study is the Automatic Dependent Surveillance-Broadcast (ADS-B) flight data. One-month ADS-B data from October 1 to October 31, 2022  have been used in this study. We get the true landing time and the landing runway of each aircraft in the dataset by matching the ADS-B positioning points to the runways of Changi airport runways. After removing the ourliers, a total of 8300 aircraft are in the dataset, and the position sampling is one location point per second. 

\subsubsection{Meteorological Data}
the meteorological data (METAR) for Singapore Changi airport station are obtained from Iowa Environmental Mesonet from Iowa State University \cite{c021}, where the wind direction (drct), wind speed in knots (sknt), visibility in miles (vsby), sky coverage (skyc1) and the altitude in feet (skyl1) are provided, and used in this study.
   
\subsubsection{Flight Plan Data}
    Flight plan (FPL) data is used to obtain the aircraft wake turbulence category (wtc). The aircraft wake turbulence are classified into four categories as Light (L), Medium (M), Heavy (Heavy), and Super (J). Our dataset contains three categories, i.e., medium, heavy, and super, except for the light aircraft.


\subsection{Data Analysis}

\subsubsection{ETA Distribution Analysis}
\begin{figure}[h]
\begin{center}
	\includegraphics[width=.956\columnwidth]{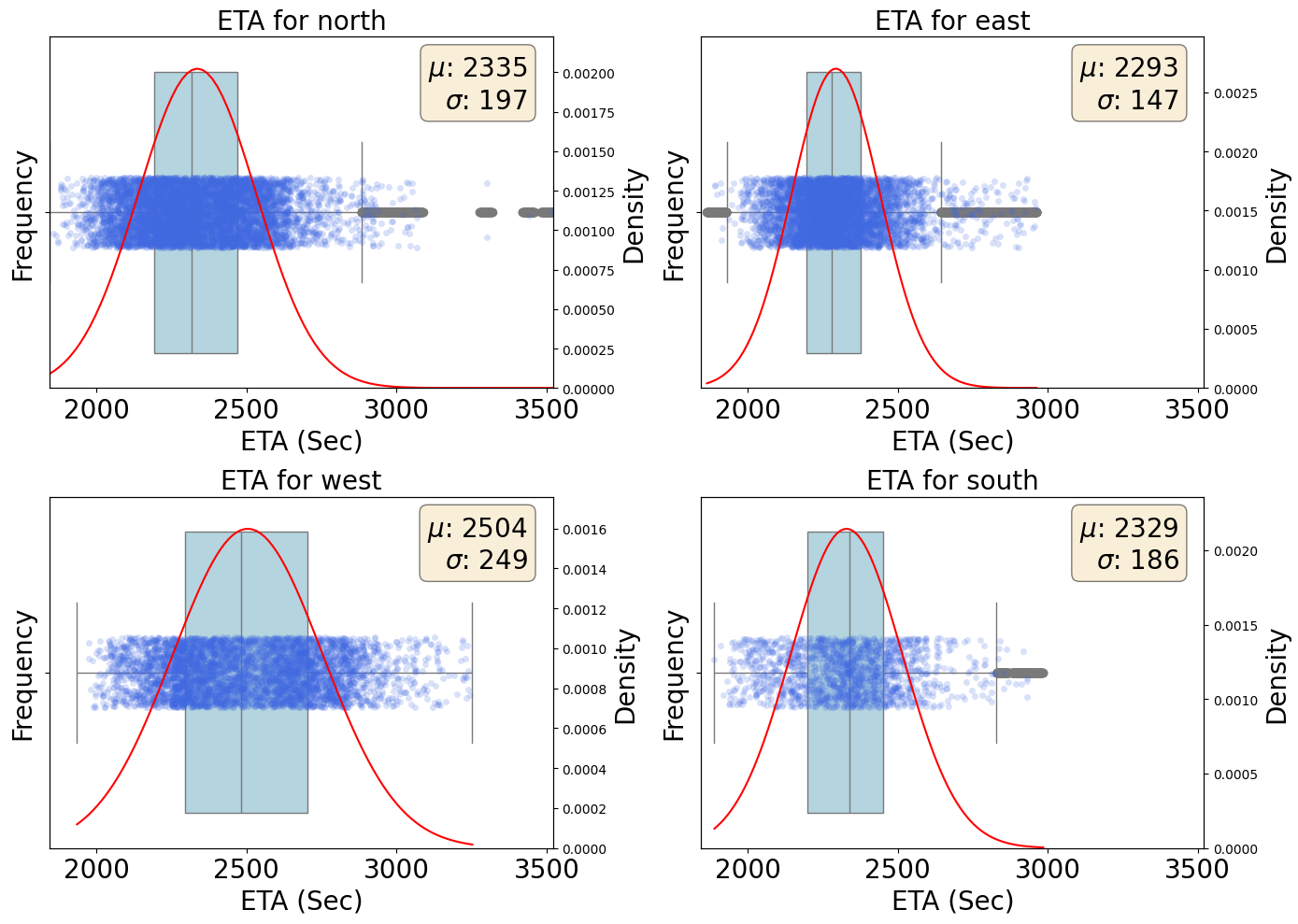}
	\caption{Box and Distribution (with Mean and Variance) for Arrival Aircraft in Four Directions at 200NM.}
	\label{eta_ana}
 \end{center}
\end{figure}

We analyze the ETA values for each arrival direction as shown in Fig. \ref{eta_ana}, where the box plot and ETA distribution for arriving aircraft from north, east, west, and south. Due to the different coverages in each direction of the trajectory data (see \ref{sub1}), we set the distance to WSSS here as 200NM. Each box represents the interquartile  range (IQR). The red line depicts the ETA distribution with the mean and variance values are shown. It shows that the range, median, mean and variance of ETA values across arrival directions differ, with aircraft arriving from west exhibiting the highest variance. This indicates the necessity of incorporating the theta degree for the airport as an input feature.

\subsubsection{ETA vs WTC}
\begin{figure}[th]
\begin{center}
	\includegraphics[width=0.9\columnwidth]{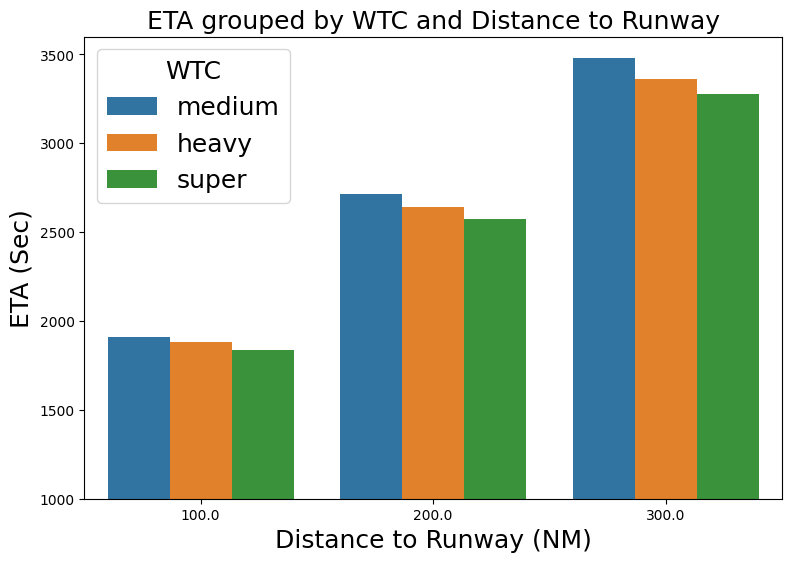}
	\caption{Mean ETA Values Regarding Aircraft Wake Turbulence Categories Given the Specific Distance to WSSS.}
	\label{eta_vs_wtc}
 \end{center}
\end{figure}

Further analysis of the aircraft wake turbulence category and the mean estimated time of arrival (ETA) is presented in Fig. \ref{eta_vs_wtc}, considering the same arrival direction and distances (100NM, 200NM, and 300NM) to WSSS. Given the arrival distance, the ETA mean value of medium aircraft is higher than the super heavy aircraft. The figure highlights the ETA gaps across different aircraft categories. Therefore, it's necessary to include the wake turbulence category as an input of the ETA modeling.

\subsubsection{ETA vs Runway Direction}

\begin{figure}[th]
   \begin{center}
	\includegraphics[width=0.95\columnwidth]{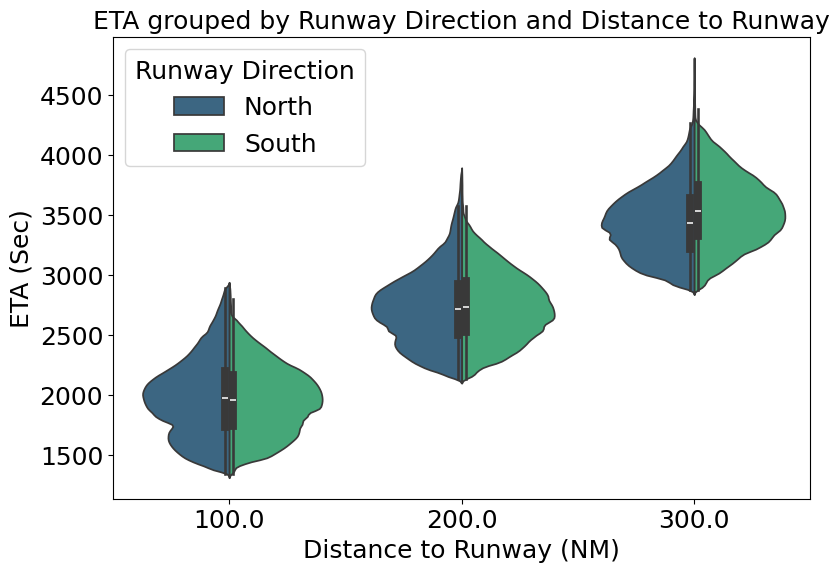}
 \caption{ETA Values with Different Runway Directions. Arrival Direction is West, and Arrival Distance is Fixed as 200NM to WSSS.}
	\label{eta_vs_runwayDrct}
 \end{center}
\end{figure}

By fixing the arrival direction as West, we analyze the runway direction (Norh landing or South landing) influence on ETA, as shown in Fig. \ref{eta_vs_runwayDrct}. The results indicate that when aircraft approach from the west, landings on the north end of the runway exhibit a bimodal distribution, whereas landings on the south end show a unimodal distribution. This difference arises because landing on the south end does not require aircraft arriving from the west to route northward, thereby reducing time consumption. Therefore, it's necessary to include the runway configuration of landing directions for the ETA modeling.

\subsection{Model Input Summary}
Based on the literature review and our data analysis, the final input features for ETA modeling in this study are presented in Table \ref{inputs}, which includes the data type and data source. Specifically, latitude, longitude, speed, and theta are derived from the ADS-B data, while day, hour are transformed from the timestamp in the ADS-B data. Here theta is calculated as the degree between based on the aircraft position and WSSS position $(y^t_{j}-y_0, x^t_j-x_0)$, where $(y^t_{j}, x^t_j)$ is the aircraft position latitude and longitude at timestamp $t$, and $(y_0, x_0)$ is the latitude and longitude of WSSS. The wake turbulence category (WTC) is obtained from the flight plan, and additional weather information is sourced from the METAR dataset. Runway direction is treated as a categorical feature. Note that the feature day, hour and wake turbulence category are set as continuous since their values have meaningful intervals, e.g., between day 1 and day 2, or hour 1 and hour 2. Similarly, the wake turbulence categories are transformed to 1, 2, 3 from medium, heavy, and super, respectively, and are also set as continuous to reflect their relative magnitudes.

\begin{table}[t]
  \begin{center}
    \caption{Input Features. cont: continuous, cat: categorical}
    \label{inputs}
    \begin{tabular}{|c|c|c|}
    \hline
      \textbf{Feature} & \textbf{Type}  &\textbf{Source} \\
      \hline
      lat, lon & \textcolor{brown}{cont}  & ADS-B\\
      \hline
      ground speed & \textcolor{brown}{cont} & \\
      \cline{1-1}
      theta &   & derived from ADS-B\\
      \cline{1-2}
      runway direction & \textcolor{green}{cat} & \\
      \hline
     day & \textcolor{brown}{cont}  & derived from ADS-B Timestamp\\
      \cline{1-1}
      hour &   & \\
      \hline
     wtc & \textcolor{brown}{cont}  & Flight Plan\\
      \hline
      vsby &   & METAR(visibility)\\
      \cline{1-1}
      skyc1 &   & METAR(sky level 1 coverage)\\
      \cline{1-1}
      skyl1 & \textcolor{brown}{cont} & METAR(sky level 1 altitude in feet)\\
      \cline{1-1}
      drct &   & METAR(wind direction)\\
      \cline{1-1}
      sknt &   & METAR(wind speed)\\
      \hline
    \end{tabular}
  \end{center}
\end{table}

\section{Experiments and Analysis}
\label{experiment}

\subsection{Experimental Settings}
We randomly split the 31 days of data into sub-datasets of 23, 4 and 4 days for model training, validation, and testing, respectively, using the input features listed in Table \ref{inputs}. The loss function is set as L1 loss, i.e., the Mean Average Error (MAE) for both XGBoost and the feature tokenization based transformer (FTT) models. The XGBoost (XGB) is configured as "gpu$\_$hist" and the predictor is set as "gp$\_$predictor". The optimizer for FTT is set as AdamW. For the FTT model, we evaluate the number of blocks and the model dimension to compare its performance with the XGBoost model. 

\subsection{Evaluation Metrics}
To measure the accuracy of the aircraft landing time predictions, we adopted the commonly used metrics MAE (Mean Absolute Error) \eqref{MAE}, MAPE (Mean Absolute Percentage Error) \eqref{MAPE}, Root Mean Squared Error (RMSE) \eqref{RMSE}. Since the ground truth values have a large range due to aircraft's distance to Changi airport, e.g., the time to landing for an aircraft at 300NM should be quite higher than the aircraft at 50NM, hence the MAPE is utilized to quantified the prediction accuracy for synchronizing distance differences. 

\begin{equation}\label{MAE}
  MAE = \frac{1}{n}\sum_{i=1}^{n}|y_i-\hat{y}_i|
\end{equation}

\begin{equation}\label{MAPE}
 MAPE = \frac{1}{n}\sum_{i=1}^{n}\frac{|y_i-\hat{y}_i|}{y_i}
\end{equation}

\begin{equation}\label{RMSE}
 RMSE = \sqrt{\frac{{\sum_{i=1}^{n}{|y_i-\hat{y}_i|}^2}}{n}}
\end{equation}

Here, $n$ is the number of test samples, $y_i$ is the ground truth label for the test sample $i$, and $\hat{y}_i$ is the corresponding estimation. For all four metrics, lower values denote better estimations. RMSE is sensitive to the outliers and MAE is actually the L1 loss that directly represents the average estimation errors. MAPE \eqref{MAPE} helps generally evaluate the estimations in terms of large ranges of ground truth values. 

\subsection{Result Analysis}
\begin{figure}[th]
\begin{center}
	\includegraphics[width=\columnwidth]{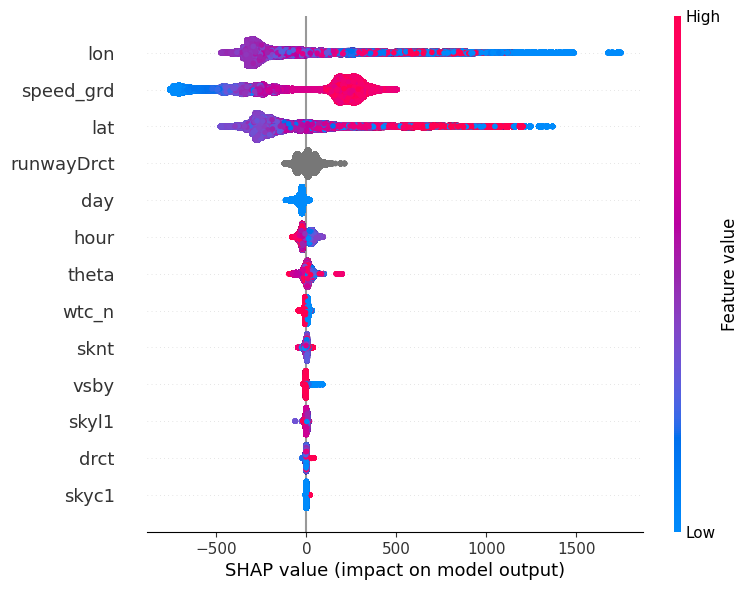}
	\caption{Feature Importance from XGBoost by shap}
	\label{feature_importance}
 \end{center}
\end{figure}

Firstly, the feature importance determined using XGBoost is illustrated in Fig. \ref{feature_importance}. The results indicate that longitude, latitude, ground speed, and the runway direction are the most significant features among all the input variables. This finding aligns with existing background knowledge, as these features are well-known to influence the performance and behavior of aircraft. Previous studies have also highlighted the critical role of geographical coordinates and temporal patterns in aviation analytics, thereby reinforcing the validity of our results.

\begin{table}[ht]
  \begin{center}
    \caption{FTT Improvement Compared to XGBoost with Same Inputs}
    \label{cmp}
    \begin{tabular}{@{}lcccr@{}} 
    \hline
      Model     & Training Time (sec) & MAE& MAPE & RMSE\\
      \hline
      \textcolor{blue}{XGBoost}      & 180              & 112.22  & 0.074 & 113\\
      \textbf{{FTT($d_{model}$=192)}}      & 67       & 110.11 & 0.072 & 110 \\
      \hline \hline
     \textbf{\textcolor{red}{FTT(($d_{model}$=256)}}      & \textcolor{red}{71 }      & \textcolor{red}{104.60} &\textcolor{red}{ 0.066} & \textcolor{red}{104}\\
      Improve (vs XGBoost)   &     60.6\%           & 7.1\%  & 10.8\% & 7.9\% \\
      \hline \hline
    \end{tabular}
  \end{center}
\end{table}

\begin{figure*}[t]
   \begin{center}
	\includegraphics[width=0.9\textwidth]{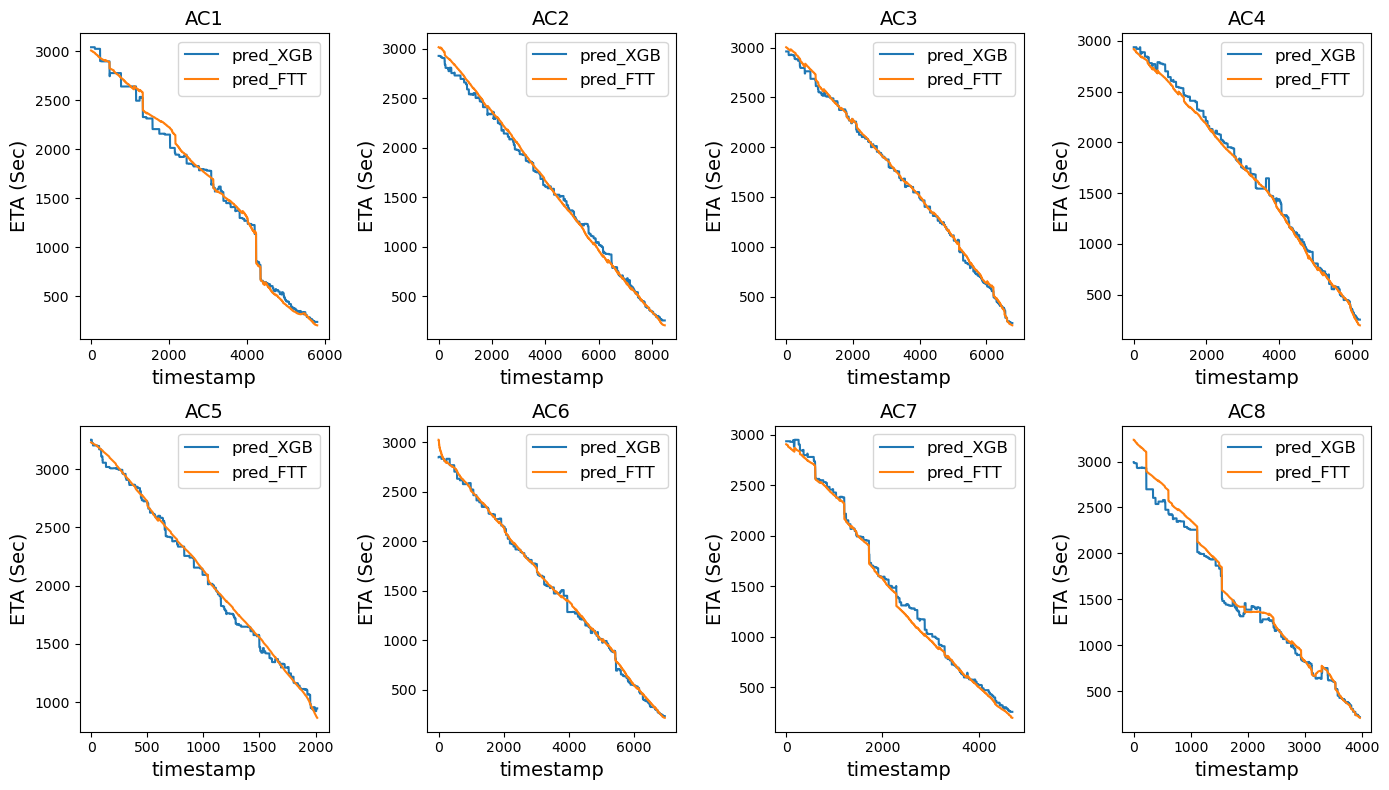}
	\caption{Samples of 1HZ Predicted ETA from XGBoost and the Feature Tokentization Transformer.} 
	\label{pred_examples}
 \end{center}
\end{figure*} 
We evaluate the performance of the FTT model performance by comparing prediction accuracy and training time while varying the model dimension (number of blocks is set as 2). The results are shown in Table \ref{cmp}.
We can observe that, when $n_{block}=2$ and $d_{model}=192$, the prediction accuracy measured by MAE shows a slight improvement over XGBoost. However, the training time for the FTT model is significantly reduced compared to the time required for training XGBoost. It's important to note that XGBoost is renowned for its efficiency due to techniques such as the gradient boosting framework and tree pruning. The FTT model is even more efficient because the Tranformer neural network uses parallel processing to calculate the projection matrices. Furthermore, when we slightly increase the model dimension to $d_{model}=256$ while keeping the $n_{block}=2$, the training time is slightly increased, but the prediction accuracy improves further. This improvement is due to the increased model dimension allowing the feature tokenizer to project the input features into a higher dimension space. The multi-head self-attention mechanism in the Transformer neural network can then better capture the import aspects of the tokenization embeddings, ultimately enhancing the model's prediction accuracy. 

We show the real-time predictions of both XGB and FTT in Fig. \ref{pred_examples}, where the prediction frequency is 1HZ. We can observe that the XGBoost predictions exhibit significant oscillations, which destabilize runway slot allocation, leading to frequent switches between adjacent aircraft. For example, if aircraft A1, A2 have ETAs of 10:05 AM and 10:08 AM at timestamp $t$, aircraft A2 is scheduled to land after A1. However, if at the next timestamp $t+1$ the predictions are changed to 10:06 AM and 10:04 AM, the sequence between this two aircraft is switched, which should be avoided in the arrival aircraft management system. In contrast, as shown in Fig. \ref{pred_examples}, the FTT predictions are smooth over time, providing more stable and reliable runway slot allocations. In a real runway slot allocation system utilizing this ETA prediction model, further post-processing is necessary to ensure the separation of landing aircraft.





\begin{figure*}[h]
   \begin{center}
	\includegraphics[width=0.8\textwidth]{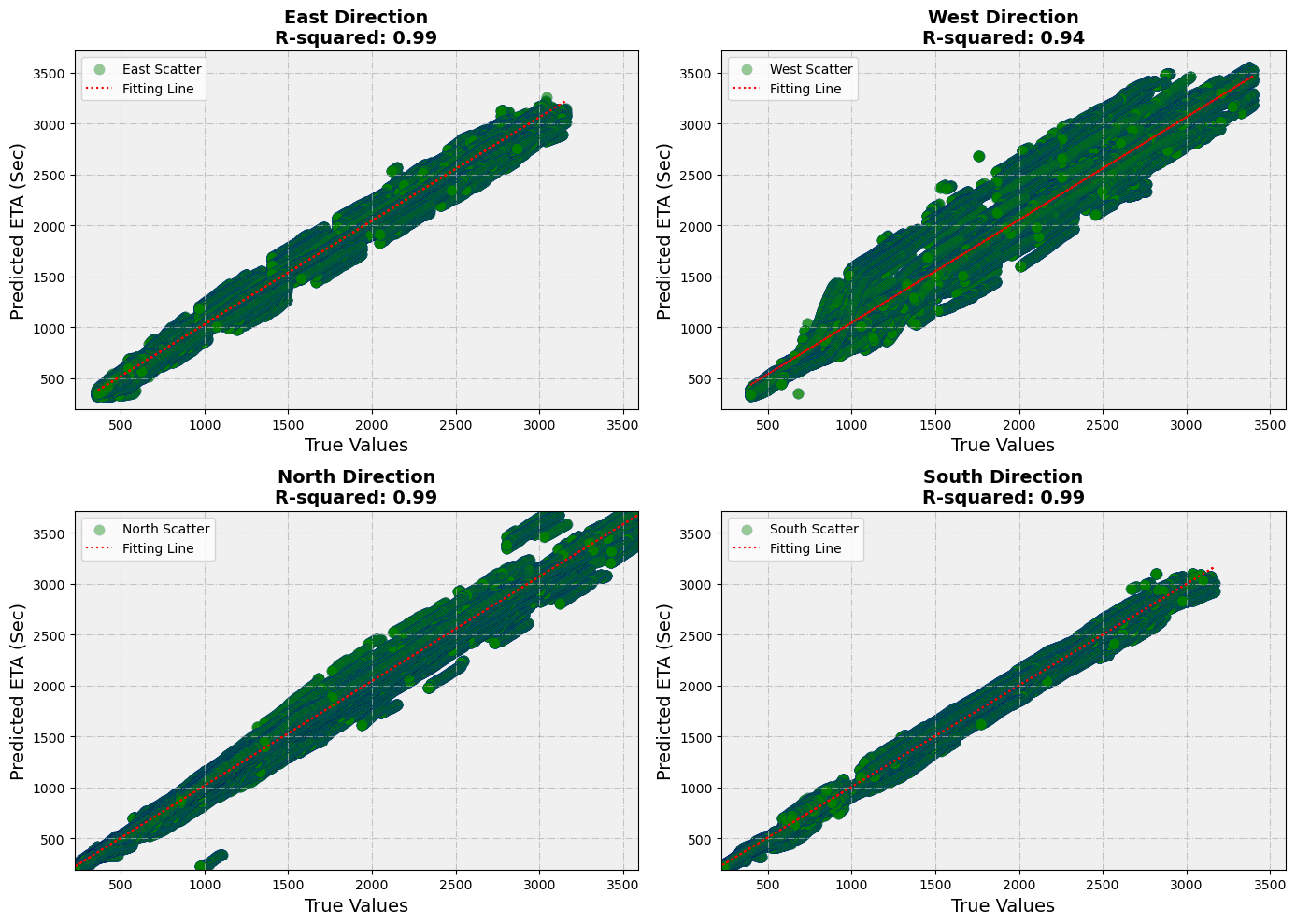}
	\caption{Scatter Plots of Predicted ETA and Ground Truth for Each Arrival Direction. }
	\label{pred_3050}
 \end{center}
\end{figure*}

We present the ETA predictions and the true ETA vales of the FTT model in Fig. \ref{pred_3050}. The results indicate that the R-Square values across all arrival directions are consistently high, reaching 0.99, except for the arrival aircraft in west (R-Square is 0.94 due to high variance of ETAs. See Fig. \ref{eta_ana}). Only a few outlier points deviate from the fitting line, denoting that the model can achieve high prediction accuracy despite varying uncertainty levels in each arrival direction.

Furthermore, we define the $BadRatio_{\gamma}$ \eqref{BadRatio} to evaluate the ratio of the bad predictions. Three evaluation metrics are described in \eqref{BadRatio}.
$BadRatio_{\gamma}$ is to evaluate the robustness of the model, where $\sigma(x)$ is a sign indicator function, i.e.,  $\sigma(x)=1$ if the expression $x$ is true, otherwise 0. $\gamma$ is a manually set threshold value, e.g., $\gamma=0.3$ represents that the absolute percentage errors $\frac{|y_i-\hat{y}_i|}{y_i}$ are higher than $30 \%$, i.e., the relative accuracy would be lower than  $70 \%$. Hereafter, the metric $BadRatio_{\gamma}$ denotes the ratio of bad predictions over the total number of test samples, Hence, a lower $BadRatio_{\gamma}$ presents a better robustness of the model.

\begin{equation}\label{BadRatio}
 BadRatio_{\gamma} = \frac{\sum_{i=1}^{n}\mathcal{\sigma}(\frac{|y_i-\hat{y}_i|}{y_i}>\gamma)}{n}
\end{equation}

The prediction accuracy of the FTT model, measured my MAE, MAPE, and BadRatio for different arrival directions and aircraft distances to WSSS is summarized in Table \ref{ftt_pred}. The results demonstrate a stable prediction accuracy across all arrival directions and aircraft positions, despite the varying control uncertainties associated with different landing times.

\begin{table*}[h]
\caption{Prediction Accuracy of the FTT Model w.r.t Arrival Direction and Distance to WSSS.}
    \label{ftt_pred}
    \centering
    \begin{tabular}{|c|c|c|c|c|c|c|c|c|c|c|c|}
 \hline

  \multirow{3}{*}{Direction}   &  \multirow{3}{*}{distance to WSSS (NM)}  & \multicolumn{2}{c}{MAE}  & \multicolumn{2}{|c}{MAPE (\%)}  & \multicolumn{2}{|c}{RMSE} & \multicolumn{4}{|c|}{BadRatio (\%)}\\
  \cline{3-12}
        &              & mean & median & mean & median & mean & median & $\gamma=0.1$  & $\gamma=0.2$  & $\gamma=0.3$ & $\gamma=0.4$  \\
\hline \hline

        &(250, 300)    & 120.40  & 85.15  &  4.06  &  3.03 & 120.40 & 85.15  &  6.89  & 1.13  &  0.17  & 0.0\\
        &(200, 250)    & 115.63  & 80.37  &  4.48  &  3.27 & 115.63 & 80.37  &  9.64  & 1.97  &  0.11  & 0.11\\
   East &(150, 200)    & 103.51  & 76.40  &  4.82  &  3.80 & 103.51 & 76.40  &  11.10  & 1.52  &  0.07  & 0.0 \\
        &(100, 150)    & 102.43  &  72.68 &  5.85  &  4.41 & 102.43 & 72.68  &  15.19  & 2.83  &  0.06  & 0.02\\
    &{(50, 100)}& 89.58  & 60.59  &  7.73  &  5.06  & 89.58 & 60.59  &  18.44  & 5.42  &  1.25  & 0.35\\
    &{(10, 50)} & 36.97  & 30.95  &  8.79  &  5.75 & 71.59 & 36.97  &  28.21  & 10.91  &  5.51  & 1.93\\ 
\hline \hline

        &(250, 300)    & 88.70  & 85.64  &  4.10  & 3.00 & 127.59 & 88.70  &  8.05  & 1.18  &  0.0  & 0.0\\
        &(200, 250)    & 87.73  & 83.17  &  4.64  & 3.45 & 124.34 & 87.73  &  9.74  & 1.31  &  0.08  & 0.0\\
  North &(150, 200)    & 85.13  & 73.52  &  5.29  & 3.95 & 120.21 & 85.13  &  12.99  & 1.64  &  0.3  & 0.0 \\
        &(100, 150)    & 79.65  & 72.10  &  6.27  & 4.62 & 117.48 & 79.65  &  18.28  & 3.49  &  1.23  & 0.05\\
        &(50, 100)     & 63.29  & 51.02  &  7.39  & 5.08 & 105.08 & 63.29  &  21.65  & 6.94  &  2.83  & 1.39\\
        &(10, 50)      & 28.09  & 19.48  &  8.17  & 4.61 & 56.63 & 28.09  &  24.46  & 10.10  &  4.85  & 2.13\\
\hline \hline

        &(250, 300)    & 95.95  & 69.80  &  4.02  &  3.55 & 113.11 & 95.95  &  5.68  &  0.005 &  0.0  & 0.0\\
        &(200, 250)    & 84.60  & 58.58  &  4.20  &  3.53 & 105.67 & 82.59  &  6.51  & 0.255  &  0.0 & 0.0\\
South   &(150, 200)    & 82.59  & 54.69  &  5.12  &  4.15 & 107.71 & 84.60  &  8.65  & 0.231  &  0.0  & 0.0\\
        &(100, 150)    & 70.72  & 53.39  &  5.79  &  4.42 & 99.89 & 70.72  &  14.28  & 1.404  &  0.05  & 0.0\\
        &(50, 100)     & 63.10  & 52.11  &  6.93  &  5.29 & 99.50 & 63.10  &  20.06  & 4.066  &  0.96  & 0.12\\
        &(10, 50)      & 24.61  & 22.18  &  7.55  &  5.75 & 38.15 & 24.61  &  24.85  & 5.634  &  1.66  & 0.57\\ 
\hline \hline

        &(250, 300)    & 124.68  & 121.09  &  5.46  &  4.21 & 167.22 &  124.68 &  14.58   & 0.97  &  0.0  & 0.0\\
        &(200, 250)    & 117.37  & 117.65  &  5.68  &  4.44 & 153.42 & 117.37  &  15.96  & 1.43 &  0.0  & 0.0\\
  West  &(150, 200)    & 104.29  & 111.09  &  6.38  &  4.63 & 150.75 & 104.29  &  18.58  & 4.77  &  0.12  & 0.0\\
        &(100, 150)    & 98.38  & 100.22  &  7.69  &  5.34  & 148.57 & 98.38  &  25.86  & 8.01  &  1.48  & 0.03\\
        &(50, 100)     & 95.14  & 81.26   &  9.67  &  6.67  & 141.88 & 95.14  &  34.70  & 11.33  &  6.33  & 1.81\\
        &(10, 50)      & 52.33  & 44.74   &  9.63  &  6.34  & 92.83 & 52.33  &  33.49 & 11.89 &  5.52  & 2.54\\ 
\hline
      
\end{tabular}
\end{table*}

Additionally, we tested the inference time consumption for one batch of aircraft, assuming 40 arriving aircraft in the airspace at a given timestamp. The result shows that it only requires 51.7 microseconds for the ETA inference for all these aircraft, demonstrating the model's efficiency. This indicates that he proposed method is capable of handling real-time ETA requests at a high frequency, e.g., 1HZ.








\section{Conclusions}
\label{conclusion}
\label{conclusion}
In this paper, we proposed utilizing a feature tokenization based Transformer for real-time aircraft ETA prediction. The tokenizer efficiently handles minimal data inputs by transforming raw data into meaningful tokens, relieving us from the complex feature engineering. The use of feature tokenization based Transformers in this context addresses the challenges of data sparsity and variability in aviation operations. Moreover, the multi-head self-attention mechanisms in the Transformer can capture complex patterns and dependencies involved in the input feature embeddings, enabling accurate predictions. Additionally, the parallel computation characteristic of the Transformer makes it computationally efficient. This type of model is capable for real-time aircraft ETA prediction due to its parallel computation ability. It can respond to the aircraft ETA request swiftly and can handle high-frequency predictions efficiently. Experimental results show that our method improve prediction accuracy by 7.1\% compared to the commonly used tree-based XGBoost model, while requiring only 39\% of its training time. It also show that the FTT model can achieve smooth predictions compared to the tree-based model, which helps stabilize the runway slot allocation in a real-time arrival management system.

\section*{Acknowledgment}

This work is supported by the National Research Foundation, Singapore, and the Civil Aviation Authority of Singapore (CAAS), under the Aviation Transformation Programme. (Grant No. ATP2.0\_WIC\_I2R for ATM\_I2R\_2). Any opinions, findings and conclusions, or recommendations expressed in this material are those of the authors and do not reflect the views of the National Research Foundation, Singapore, and the Civil Aviation Authority of Singapore. The authors would like to thank all colleagues from CAAS for providing valuable comments and suggestions on this work.

\vspace{12pt}

\end{document}